\documentclass[10pt,twocolumn,letterpaper]{article}

\usepackage{wacv}              

\usepackage{graphicx}
\usepackage{amsmath}
\usepackage{amssymb}
\usepackage{booktabs}
\usepackage{subcaption} 
\usepackage{url}

\usepackage[pagebackref,breaklinks,colorlinks]{hyperref}

\usepackage[capitalize]{cleveref}
\crefname{section}{Sec.}{Secs.}
\Crefname{section}{Section}{Sections}
\Crefname{table}{Table}{Tables}
\crefname{table}{Tab.}{Tabs.}

\def\wacvPaperID{1823} 
\def\confName{WACV}
\def\confYear{2025}

\begin{document}

\title{MVMD: A Multi-View Approach for Enhanced Mirror Detection }

\author{Yidan Shen$^{*}$, Yu Wen$^{*}$, Chen Zhang, Xin Fu, Renjie Hu\\
University of Houston\\
{\tt\small\{yshen20, ywen8\}@uh.edu, chenzh0220@outlook.com, \{xfu8, rhu7\}@central.uh.edu}
}
\maketitle

\begingroup
\renewcommand\thefootnote{*}
\footnotetext{Co-first authors} 
\endgroup

\begin{abstract}
In 3D reconstruction, mirrors introduce significant challenges by creating distorted and fragmented spaces, resulting in inaccurate and unreliable 3D models. As 3D reconstruction typically relies on multi-view images to capture different perspectives of a scene, detecting and labeling mirrors in multi-view images before reconstruction can effectively address this issue. However, existing methods focus solely on single-image detection, overlooking the rich information provided by multi-view setups. To overcome this limitation, we propose MVMD, a novel Multi-View Mirror Detection method, along with the first database specifically designed for mirror detection in multi-view scenes.

The design of MVMD is grounded in the inherent associations between objects seen from different views and those reflected inside and outside of mirrors. These relationships are learned through cross- and self-attention mechanisms. MVMD consists of three key blocks: the Inter-Views Block tracks the shifts of objects within mirrors caused by changes in viewpoint; the Intra-View Block detects object reflections inside mirrors; and the Refinement Block sharpens mirror boundaries and enhances detected details.

Experimental results show that our method improves accuracy by up to 2.6\% and IoU by up to 11.1\%, compared to single-image mirror detection techniques. This substantial improvement makes MVMD particularly effective for computer vision tasks, especially in enhancing the accuracy of 3D reconstruction in mirror-dense environments. Code and data are available at: https://github.com/mvmdwacv25.

\end{abstract}

\section{Introduction}
\label{sec:intro}

Mirrors pose a major challenge to 3D scene reconstruction, as they mislead algorithms into interpreting reflections as real objects, resulting in incorrect depth estimates and a distorted spatial structure. This misinterpretation complicates the accurate reconstruction of a scene, with reflections often appearing as extensions of the real environment, leading to errors in object recognition and scene analysis. Traditional 3D reconstruction methods and state-of-the-art algorithms, including NeRF \cite{mildenhall2021nerf}, 3DGS \cite{kerbl3Dgaussians}, and Multi-view Stereo Vision (MVS) models (e.g., COLMAP \cite{schoenberger2016mvs}), all struggle with mirror-related issues. Mirrors introduce problems such as phantom objects \cite{meng2024mirror} and distorted geometries \cite{meng2024mirror, zeng2023mirror-nerf}, making it challenging for these techniques to accurately differentiate between real objects and mirror reflections, resulting in poor reconstruction quality in mirror-rich areas.



Identifying mirrors before 3D reconstruction can significantly enhance reconstruction performance. However, current mirror detection algorithms \cite{mirrornet, pdnet, PMD, sanet, satnet, ardent-s} are primarily designed for single-view scenarios, limiting their effectiveness in multi-view 3D reconstruction, where consistent mirror identification from different angles is crucial. Although video-based methods \cite{vmd, Warren_2024_CVPR, xu2024zoom} incorporate temporal information, they often lack true multi-view perspectives due to limited camera angles, which restricts their applicability in capturing diverse spatial details required for accurate 3D mirror detection. Specifically, single-view and video methods lack the rich spatial information inherent in multi-view setups, hindering their ability to accurately detect mirrors and differentiate reflections from real objects. As a result, relying on these methods can lead to incomplete or distorted reconstructions \cite{pdnet}. While depth maps can sometimes improve single-view mirror detection by providing additional guidance \cite{pdnet, ardent-s}, they are costly to obtain and are often unavailable in many 3D reconstruction tasks. Furthermore, due to limitations in network architectures optimized for single inputs, most existing algorithms cannot effectively process multi-view stereo (MVS) images or video sequences with varying perspectives \cite{vmd, xu2024zoom, Warren_2024_CVPR}, despite their common use in 3D scene reconstruction. Therefore, designing a network that can handle multi-view inputs using only RGB images is highly desirable, as it aligns well with the data characteristics of 3D reconstruction and addresses the unique challenges of mirror detection in such scenarios.


However, developing a Multi-View Mirror Detection method using only RGB images presents several challenges. First, there is a significant lack of multi-view datasets that include mirrors, which severely limits the ability to train models effectively. Second, different viewpoints result in varied visual appearances of a scene, especially in the presence of mirrors. As the viewpoint shifts, the scene inside the mirror changes at a different rate compared to the outside. This discrepancy makes it challenging for algorithms to distinguish between changes caused by actual viewpoint shifts and those caused by mirror reflections. This is because both can produce similar visual alterations across different views. This ambiguity complicates the detection process and may lead to inaccurate scene interpretations. Finally, objects such as windows and doors can exhibit depth discontinuities in multi-view inputs similar to those caused by mirrors, further complicating the detection process \cite{ardent-s}.

To address these challenges, we propose a mirror detection method designed for multi-view RGB image inputs without relying on explicit depth information. Our approach comprises three main components. The Inter-Views Block targets changes in different views caused by mirror reflections, distinguishing them from those due to actual viewpoint shifts. It employs cross-attention and self-attention mechanisms to capture reflection movements across views. The Intra-view Block focuses on objects with depth discontinuities by conducting cross-attention between the image and its mirror-flipped version, capturing the relationships between objects inside and outside mirrors. Lastly, the Refinement Block improves the final prediction using an edge-enhancement network. To train our network, we developed a new multi-view dataset featuring 98 scenes and a total of 3,181 images. This dataset is the first to include mirrors in multi-view scenes, offering a diverse representation of mirrors across various scenarios. It serves as a valuable resource for both training and evaluating mirror detection models. Our experiments demonstrate that our method significantly outperforms state-of-the-art techniques in terms of both efficiency and accuracy. To summarize, our key contributions are as follows:

\begin{itemize}
    \item As no existing datasets meet the needs of Multi-View Mirror Detection, we have created a pioneering dataset designed to fulfill the requirements, which includes a wide range of scenes and provides an extensive collection of multi-view images.
    \item Unlike conventional methods that rely on single images or video inputs, we introduce a multi-input architecture that leverages three inputs to enhance mirror detection in multi-view applications.
    \item We implement an Inter-views Block that employs cross-attention and self-attention mechanisms to target changes caused by mirror reflections across different views, distinguishing these from those due to actual viewpoint shifts.
    \item To identify and correlate objects inside and outside a mirror, we design an Intra-view Block to address the challenge of reflection symmetry by leveraging the observation that mirror reflections are flipped versions of real objects.
    \item Extensive evaluations demonstrate that our model significantly improves IOU by up to 11.1\% compared to existing state-of-the-art methods, particularly in complex scenes with diverse object geometries and occlusions caused by objects in front of mirrors.
    
\end{itemize}




\section{Related Work}
\subsection{Semantic Segmentation}
Semantic segmentation involves classifying each pixel in an image into a specific semantic category. Traditional methods for semantic segmentation have been applied to mirror detection tasks, aiming to accurately identify mirror regions versus other objects in the scene. Recent advancements \cite{fu2019dual, chen2017deeplab, huang2019ccnet, ding2018context, chen2020bi, 10675013} often rely on single RGB images, which lack the comprehensive contextual understanding of the entire scene. Other methods incorporate depth information \cite{9774020, cheng2017locality} as an additional supervisory signal to enhance results; however, this does not apply to 3D reconstruction datasets, which often lack depth information. Moreover, many recent approaches utilize complex networks, such as fully convolution networks (FCNs) \cite{long2015fully} and pyramid modules \cite{zhao2017pyramid, chen2017deeplab}, which greatly impact network efficiency. Despite these efforts, treating mirrors as an additional semantic category with existing semantic segmentation methods often yields unsatisfactory results because these methods struggle to accurately classify the complex and varied content reflected within the mirror.

\begin{figure*}[t]
    \begin{subfigure}[b]{0.74\linewidth}
    \centering
        \includegraphics[width=\linewidth]{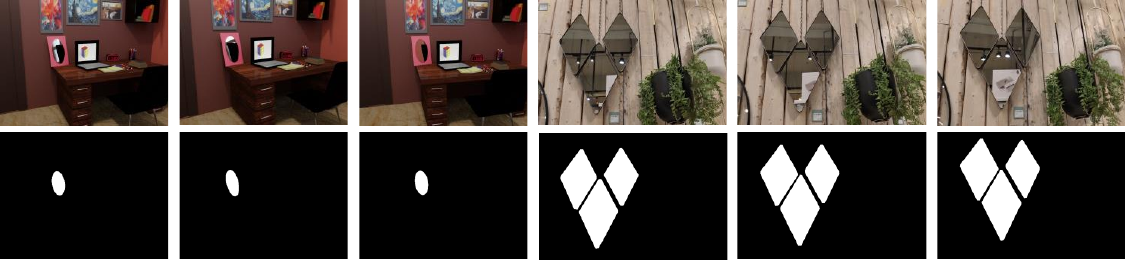}
        \caption{Illustrative examples in Multi-View Mirror Detection (MVMD) dataset.}
        \label{fig:dataset}
    \end{subfigure}
    \hspace{0.02\linewidth}  
    \begin{subfigure}[b]{0.227\linewidth}
    \centering
        \includegraphics[width=\linewidth]{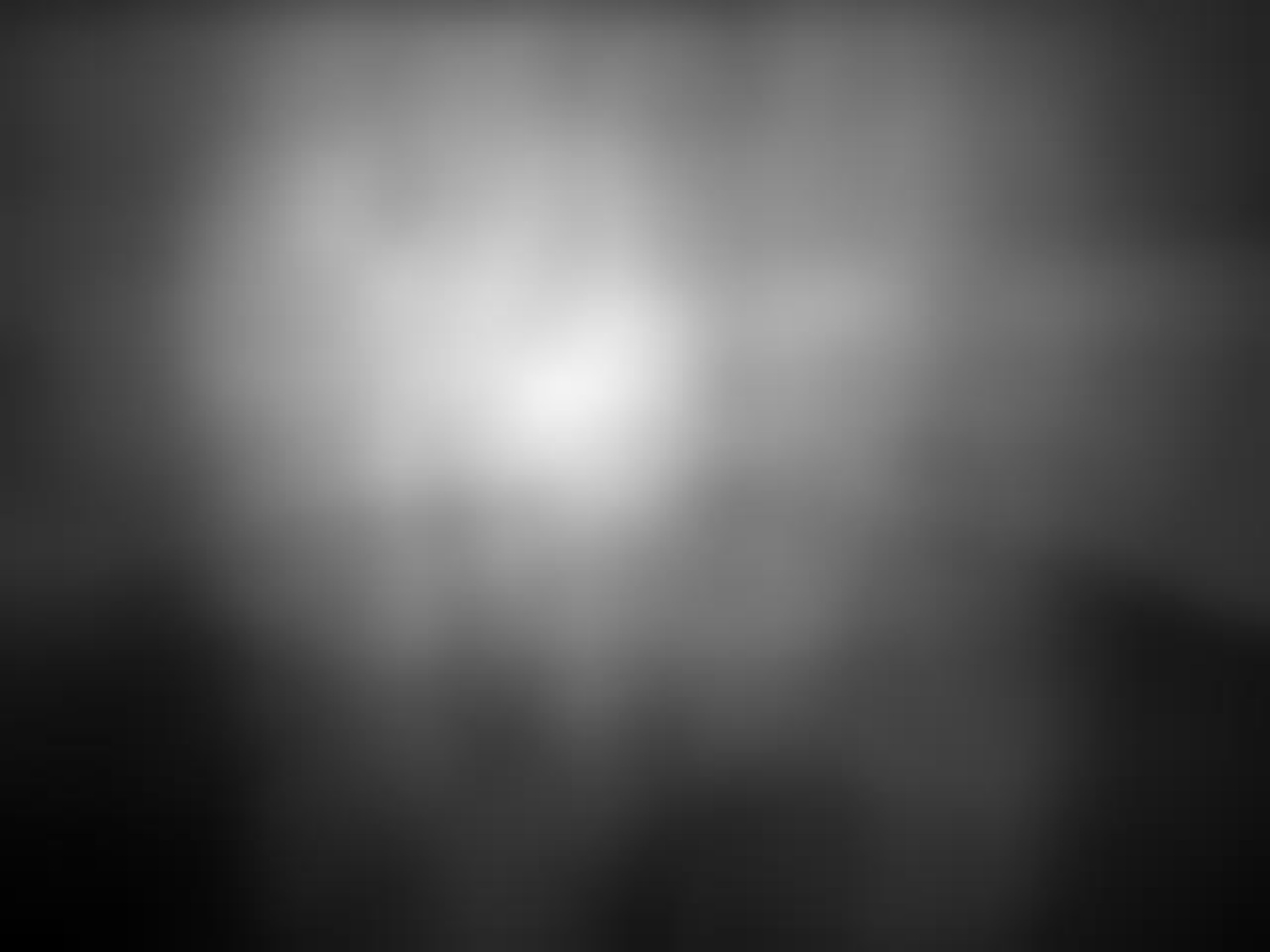}
        \caption{Spatial distribution of mirrors.}
        \label{fig: location distribution}
    \end{subfigure}
    \vspace{-0.5cm}
    \caption{Visualization and analysis of the Multi-View Mirror Detection (MVMD) dataset.}
    \vspace{-0.1cm}
\end{figure*}


\subsection{Mirror detection}
Mirror detection approaches can be broadly categorized into single-image and video-based methods, each addressing unique challenges and use cases. Single-image methods rely on static images for mirror detection, employing techniques such as edge detection and feature matching. In contrast, video-based methods utilize sequences of frames to leverage temporal information, incorporating techniques like optical flow and motion tracking.

\subsubsection{Single Image-based Mirror Detection}
Due to their simplicity, single-image-based methods have become popular in applications where computational efficiency is crucial. Yang et al. \cite{mirrornet} first utilized contextual contrast cues in a single RGB image for mirror segmentation. Mei et al. \cite{pdnet} leveraged ground truth depth to predict mirrors in a single image but relied heavily on salient discontinuities in the depth map. Similarly, Zhou et al. \cite{ardent-s} employed a student model trained with distilled knowledge and multi-modal feature fusion from the single RGB image and its corresponding depth map, also depending on salient discontinuities. Lin et al. \cite{vmd} detected mirrors by contrasting contextual relationships inside and outside their boundaries but required salient boundaries and corresponding objects. Guan et al. \cite{sanet} first acquired class-specific knowledge and then used semantic associations between mirrors and their surrounding objects to locate mirrors in a single image. Tan et al. \cite{vcnet} embedded visual charity features into the detection model, depending on the characteristics of objects inside the mirror. He et al. \cite{he2022efficientmirrordetectionmultilevel} used a Multi-Level Heterogeneous Contrasted Module and a Reflection Semantic Logical Module for accurate mirror detection. In spite of the improvements, most existing methods use complex pyramid structures that are computationally intensive and impractical for real-time applications. Moreover, these methods often struggle to distinguish mirrors from other objects such as windows, doors, and art frames. It may also perform poorly with small mirrors due to insufficient information. In contrast, our proposed method utilizes three images as input, overcoming the limitations of single-image approaches and addressing these issues without using complex structures.


\subsubsection{Video-based Mirror Detection}
Single-image-based methods cannot fully address the challenges in dynamic and complex environments, leading to the development of video-based methods that leverage temporal information for consistency between frames. Lin et al. \cite{vmd} use intra-frame and inter-frame correspondences from three video frames to predict mirror masks. However, their approach may fail when intra-frame correspondences are absent (e.g. when objects visible in the mirror are not present outside the mirror). Warren et al. \cite{Warren_2024_CVPR} enhance predictions by leveraging motion inconsistencies in optical flow fields. However, their method faces challenges in specific scenarios. It struggles when mirror reflections move at speeds comparable to other objects. Additionally, scenes with depth discontinuities, such as windows and doors, also present issues due to similar motion inconsistencies. Xu et al. \cite{xu2024zoom} introduce temporal consistency to learn from weak frame-level indicators for detecting mirrors in motion. Nonetheless, this method performs poorly in static mirror scenes, which hinders its application in 3D reconstruction.


\section{MVMD Dataset}
To advance the field of Multi-View Mirror Detection and support the training and testing of our proposed MVMD network, we present the MVMD (Multi-View Mirror Detection) dataset, filling a critical gap as the first dataset tailored for multi-view mirror detection. This comprehensive dataset comprises 98 diverse scenes and 3,181 images, specifically designed to address the challenges associated with mirror detection in multi-view environments.

Specifically, we create the MVMD dataset using Blender 4.0.1 \cite{blender.org}, a photorealistic 3D creation suite, and incorporate selected real-world scenarios from VMD \cite{vmd} and Mirror-NeRF \cite{zeng2023mirror-nerf} to ensure a diverse and realistic representation of scenes and mirrors suitable for real-world applications. We applied data augmentation techniques such as cropping and rotation to the original images to expand the dataset. Specifically, the MVMD dataset includes 1,559 images generated using Blender, with a resolution of 640×480, 1,192 images from the VMD-D dataset at a resolution of 1280×720, and 430 images from the Mirror-NeRF dataset, with resolutions of 800×800 and 400×300. Within the dataset, images from VMD-D and Mirror-NeRF were selected to ensure that inter-view angular separations exceed 0.8 degrees for adequate view variation and that each scene includes at least three multi-view images to satisfy MVMD’s input requirements. This diverse dataset provides a solid foundation for developing and evaluating Multi-View Mirror Detection methods. Sample images from the MVMD dataset are shown in \cref{fig:dataset}. For additional dataset examples, please refer to Appendix A.3.

To capture real-world diversity in mirror detection, the dataset features mirrors positioned across various regions of the images, ranging from the center to the edges. The spatial distribution of mirrors is illustrated in \cref{fig: location distribution}. Additionally, the dataset includes mirrors of different shapes and sizes, ranging from common forms like round, ellipsoid, and rectangular to less common shapes such as irregular polygons and asymmetrical forms. This variety enhances the dataset's robustness, supporting comprehensive training and testing of models for real-world applications.

 

\begin{figure*}[t]
    \centering
    \includegraphics[width=1\linewidth]{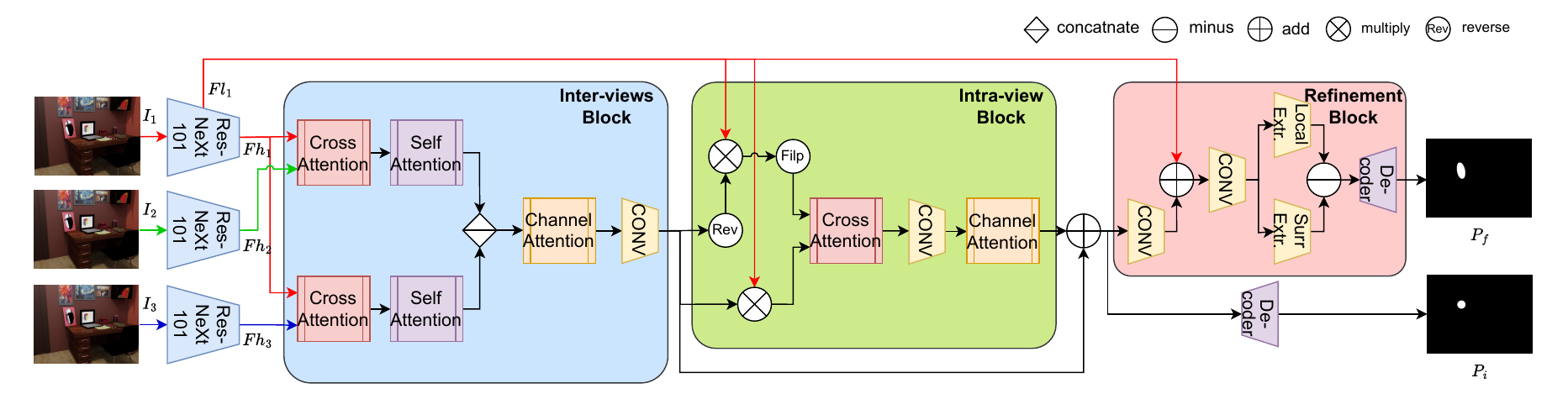}
    \caption{Overall architecture of the MVMD network. The network consists of three blocks: the Inter-Views Block, the Intra-View Block, and the Refinement Block. Information flow is depicted using colored arrows: red for $I_1$, green for $I_2$, and blue for $I_3$. The combined information flow is indicated by black arrows.}
    \label{fig:Architecture}
\end{figure*}

\section{Methodology}
\subsection{Key Insights for Multi-View Mirror Detection}
Understanding the core behavior of mirrors from varying viewpoints is crucial for designing an effective mirror detection network. This deep understanding reveals several key insights that guide the development of our proposed method, including dynamic reflections, object correspondence, and distinct edges. These insights directly inform the design of each component of our network, ensuring that our approach effectively addresses the unique challenges of mirror detection in multi-view settings.

First, as the viewpoint shifts, reflections in mirrors dynamically alter the perceived positions, orientations, and visibility of objects within the scene. Unlike objects outside the mirror, which maintain a consistent spatial relationship with the camera, reflections within the mirror behave unpredictably. The mirror's transformative effect can cause these reflections to flip and distort the scene in ways that differ from the direct view. This variability in reflections provides valuable information for detecting mirrors and guides the design of our Inter-Views Block, which is tailored to capture these dynamic differences as the viewpoint shifts.

Second, objects can appear both in actual space and as reflections on a mirror's surface, maintaining their physical properties and spatial relationships. This correspondence can be leveraged to identify the presence of mirrors. To utilize this, our Intra-View Block is specifically designed to identify and exploit these relationships, enhancing mirror detection based on object correspondences.

Finally, mirrors often exhibit distinct edges or frames that serve as clear indicators of their presence, providing additional cues for accurate detection. Our Refinement Block focuses on these features to improve detection accuracy by emphasizing the unique edges or frames of the mirrors.

\subsection{Architecture Overview}
Based on the discussion in Section 4.1, we construct the overall MVMD network, integrating the Inter-Views, Intra-View, and Refinement Blocks, as illustrated in \cref{fig:Architecture}.

Specifically, the MVMD network processes three input images, $I_1$, $I_2$, and $I_3$, captured from different viewpoints. The angle between the viewpoints of $I_1$ and $I_2$ is small, while that between $I_1$ and $I_3$ is larger. Each RGB image, with dimensions (H $\times$ W), is first fed into a pre-trained ResNeXt-101 backbone network \cite{Xie2016} to extract multi-scale features. The high-level features (from the 5th scale of the ResNeXt network) of $I_1$, $I_2$, and $I_3$ are then processed by the Inter-Views Block to identify potential mirror locations by analyzing information from these different viewpoints. Subsequently, the Intra-View Block uses the low-level feature (from the 2nd scale of the ResNeXt network) $Fl_{1}$ from $I_1$ and the output of the Inter-Views Block, $F_{Inter}$, to identify possible mirror regions by examining the relationships between objects inside and outside the mirror. The outputs of both blocks, $F_{Inter}$ and $F_{Intra}$, are combined and passed through a decoder to generate an initial mask $P_i$. Finally, this initial mask $P_i$, along with the low-level feature $Fl_{1}$, is processed through the Refinement Block. This block uses edge information to produce the final mask $P_f$. In the following sections, we will provide a detailed description of each block's functionality and how they contribute to the overall network architecture.




\subsection{Inter-views Block}
As illustrated in \cref{fig:Architecture} (blue block), the Inter-Views Block is designed to capture and analyze differences in objects within the mirror area between two images as the viewpoint changes. It employs cross-attention mechanisms to detect these differences using high-level features, while self-attention provides a comprehensive understanding of potential mirror regions.


To detect mirror reflections across different views, cross-attention is applied between the feature sets $[Fh_{1}, Fh_{2}]$ and $[Fh_{1}, Fh_{3}]$. This dual-view approach increases the likelihood of identifying reflection changes, addressing cases where a single view may not capture significant variations. The cross-attention mechanism specifically targets changes due to mirror reflections rather than direct viewpoint shifts. Each cross-attention operation is followed by a self-attention step, which focuses on interpreting the entire image and distinguishing between mirror and non-mirror areas. The attention mechanism is formulated as:
\begin{equation}
  F_{ATT} = \text{cat}\left(\mathcal{S}(\mathcal{C}\left( Fh_{1}, Fh_{2}\right)), \mathcal{S}(\mathcal{C}\left( Fh_{1}, Fh_{3}\right))\right),
  \label{eq:relation2}
\end{equation}
where $\mathcal{C}(.,.)$ represents cross-attention, $\mathcal{S}(.)$ denotes self-attention, $\text{cat}(.,.)$ indicates concatenation along feature channels, and $F_{ATT}$ is the resulting output.



To enhance feature representation, we use a channel attention mechanism inspired by \cite{woo2018cbamconvolutionalblockattention}. This involves max and average pooling to obtain channel descriptors, which are processed through a shared MLP. The weighted features are then integrated using a convolution network with Batch Normalization and ReLU activation. The Inter-Views Block's process is summarized by:
\begin{equation}
 F_{Inter} = \mathcal{CONV}\left(\mathcal{CA} \left( F_{ATT} \right)\right),
  \label{eq:relation1}
\end{equation}
where $\mathcal{CA}$ denotes channel attention, $\mathcal{CONV}$ represents convolution, and $F_{Inter}$ is the output of the Inter-View Block. The input and output of the Inter-View Block have dimensions $(H \times W)/16$ and 256 channels.


\subsection{Intra-view Block}

As illustrated in \cref{fig:Architecture} (green block), the output from the Inter-Views Block is integrated with the target image information in the Intra-View Block to capture relationships between corresponding objects inside and outside the mirror. Unlike existing methods that rely only on flipped versions or mask guidance, our approach incorporates cross-attention mechanisms between the mirror and non-mirror areas of the target image, utilizing horizontally flipped versions for non-mirror regions. This design enables the network to learn correspondences accurately and efficiently.

In the Intra-View Block, the output from the Inter-Views Block, $F_{Inter}$, is multiplied by the target image's low-level feature, $Fl_{1}$, to isolate the mirror area:
\begin{equation}
    F_{Mir} = Fl_{1} \times F_{Inter},
\end{equation}
where $F_{Mir}$ represents the feature map highlighting the mirror area. Conversely, to isolate the non-mirror area, $F_{Inter}$ is subtracted from 1 and multiplied with $Fl_{1}$, followed by a horizontal flip:
\begin{equation}
    F_{NoMir} = \mathcal{FLIP}(Fl_{1} \times (1 - F_{Inter})),
\end{equation}
where $F_{NoMir}$ denotes the flipped feature map of the non-mirror area, and $\mathcal{FLIP}$ represents the horizontal flip operation.

These two feature maps, $F_{Mir}$ and $F_{NoMir}$, are then input into a cross-attention operation to learn the association between objects inside and outside the mirror. The cross-attention ensures accurate correspondence by aligning both mirror and non-mirror features to the same orientation. Additionally, channel attention is employed to guide the network's focus toward crucial information. The process of the Intra-View Block can be summarized as follows:
\begin{equation}
    F_{Intra} = \mathcal{CA}\left(\mathcal{CONV}(\mathcal{C}(F_{Mir}, F_{NoMir}))\right),
\end{equation}
where $\mathcal{CA}$ denotes channel attention, $\mathcal{CONV}$ represents the convolution operation, and $\mathcal{C}$ stands for cross-attention. The input features $Fl_{1}$ and $F_{Inter}$ have dimensions $(H \times W)/8$ and $(H \times W)/16$, respectively, with 256 channels. The output, $F_{Intra}$, retains dimensions $(H \times W)/16$ and 256 channels.

\subsection{Refinement Block}
By directly integrating the outputs from the Inter-Views and Intra-View Blocks, we generate an initial mirror mask that effectively covers the mirror areas. The process can be formulated as:
\begin{equation}
    P_{c} = \mathcal{D}( F_{Intra} +  F_{{Inter}}),
\end{equation}
where $\mathcal{D}$ is the decoder that produces the initial mask $P_{i}$.

Although we have obtained an initial mask that effectively covers the mirror areas and aligns with the intended design, it still suffers from rough edges and noise, which limits its applicability in scenarios requiring high precision. To address these issues, we design the Refinement Block for edge enhancement and noise reduction. It processes the combined features from the Inter-Views Block ($F_{Inter}$) and the Intra-View Block ($F_{Intra}$), along with the low-level features $Fl_{1}$ of the target image, as follows:
\begin{equation}
    F_f = \mathcal{CONV}( F_{Intra} +  F_{{Inter}})+Fl_{1}.
\end{equation}
The edge-enhancement network applies two parallel convolution layers to the combined input $F_{f}$: one for local feature extraction and another for surrounding feature extraction. The local extractor focuses on fine-grained details within small regions using a 3×3 kernel and a dilation rate of 1. In contrast, the surrounding extractor captures broader contextual information using a 5×5 kernel and a dilation rate of 2. The outputs of these two extractors are then subtracted to isolate edges by emphasizing the differences between local features and broader context. A decoder with convolution layers processes the result to produce the final mask $P_{f}$. This process is mathematically formulated as:
\begin{equation}
    P_{f} = \mathcal{D}\left(\mathcal{E_L}(\mathcal{CONV}(F_f)) - \mathcal{E_S}(\mathcal{CONV}(F_f) )\right),
\end{equation}
where $\mathcal{E_L}$ denotes the local feature extractor, $\mathcal{E_S}$ denotes the surrounding feature extractor, and $\mathcal{D}$ represents the decoder that produces the final binary mask $P_{f}$. The block takes feature maps with dimensions $(H \times W)/8$ and $(H \times W)/16$, each with 256 channels, and outputs a binary mask of size $H \times W$.

\subsection{Loss Function}
We use Mean Squared Error (MSE) as the loss function to train the MVMD network because it effectively measures the pixel-wise discrepancies between the ground truth masks $G$ and the predicted masks $P_i$ of mirrors. The loss function has two main components. The first component, $L(P_c, G)$, measures the error of the initial mask $P_i$ and helps train the Inter-Views Block and Intra-View Block. The second component, $L(P_f, G)$, evaluates the error of the final mask $P_f$ and is crucial for training the entire network.

To ensure the initial mask effectively guides the Inter-Views and Intra-View Blocks and that the final mask loss properly supervises the overall network performance, we assign a weight factor of 2 to the loss associated with the final mask. This weighting balances the contributions of both loss components while avoiding excessive penalization of the initial mask. Thus, the total loss is computed as:
\begin{equation}
    L_{\text{Total}} = L(P_c, G) + 2 \times L(P_f, G),
  \label{eq:loss}
\end{equation}
where $L(.,.)$ represents the MSE calculation and $G$ is the ground truth mirror mask for the target image $I_1$.





\begin{figure*}[t]
    \centering
    \includegraphics[width=1\linewidth]{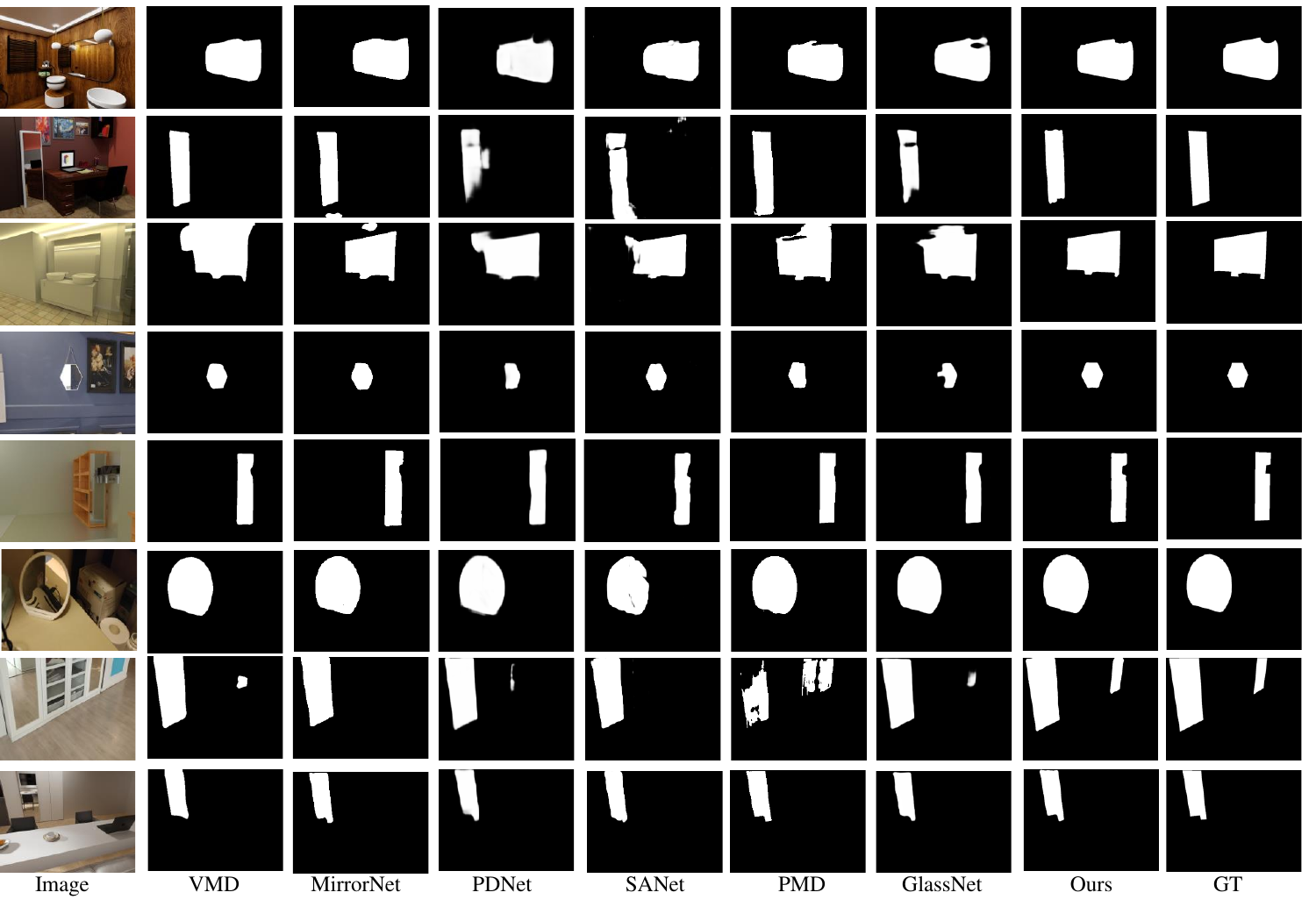}
    \vspace{-0.4cm}
    \caption{Visual comparisons of our method with state-of-the-art methods.}
    \label{fig:SOTAcompare}
    \vspace{-0.2cm}
\end{figure*}

\section{Experiment}
\subsection{Dataset and Implementation}
From the proposed MVMD dataset, we use 2,798 images from 85 scenes (87\%) for training and 383 images from 13 scenes (13\%) for testing and validation. The dataset is split using random sampling to ensure that the model is evaluated in diverse environments, thereby simulating real-world performance. We apply the hold-out validation strategy to ensure an unbiased evaluation. For uniform processing, all images and their corresponding masks are resized to 640 $\times$ 480.

The MVMD network is implemented in PyTorch and trained on a cluster with an A100 GPU. We use ResNeXt-101 \cite{Xie2016}, pre-trained on ImageNet, as the backbone network to extract image features. The training utilizes the Adam optimizer with a batch size of 2, momentum of 0.9, weight decay of $5 \times 10^{-4}$, and a base learning rate of 0.0001. We apply cosine learning rate decay with a 3-epoch warm-up period to adjust the learning rate over 20 epochs.



\begin{table}[t]
\centering
\begin{tabular}{l c c c c}
\hline
Method & IOU& MAE& Accuracy& NMSE\\
\hline
VMD\cite{vmd}& 0.8808& 0.0150& 0.9850& 0.1658\\
MirrorNet\cite{mirrornet}& 0.8794& 0.0177& 0.9823& 0.1707\\
PDNet\cite{pdnet}& 0.8181& 0.0184& 0.9816& 0.2276\\
SANet\cite{sanet}& 0.7903& 0.0362& 0.9638& 0.3469\\
PMD\cite{PMD}& 0.8823& 0.0165& 0.9835& 0.1574\\
GlassNet\cite{glassnet}& 0.8603& 0.0163& 0.9837& 0.1688\\
Ours & \textbf{0.9019}& \textbf{0.0106}& \textbf{0.9894}& \textbf{0.1183}\\
\hline
\end{tabular}
\caption{Comparison of state-of-the-art methods. The best results are highlighted in bold for reference.}
\vspace{-0.2cm}
\label{tab:SOTA Comparison}
\end{table}

\subsection{Results Analysis}
\textbf{Detection Accuracy}
Given the absence of established methods for Multi-View Mirror Detection, we evaluate our approach by comparing it to six state-of-the-art techniques from related fields: VMD \cite{vmd} for video mirror detection, MirrorNet \cite{mirrornet}, PMD \cite{PMD}, PDNet \cite{pdnet}, and SANet \cite{sanet} for single-image mirror detection, and GlassNet \cite{glassnet} for glass surface detection. All methods are re-trained on our dataset. For PDNet, which requires ground truth depth, we generate depth using the state-of-the-art single-image depth estimation method \cite{depthanything}. For single-image mirror detection, we treat each view as an independent image for prediction. As shown in \cref{tab:SOTA Comparison}, our method outperforms all competitors by a significant margin across all four metrics, with a notable 11.1\% improvement in IOU and a 2.6\% increase in accuracy, leveraging angle-diverse images for enhanced mirror detection.


\textbf{Network Size}
\cref{tab:Complexity comparison} provides a quantitative assessment of MVMD compared to several state-of-the-art methods, which includes metrics such as parameter count, memory usage, and $F_{\beta}$ score to illustrate the computational complexity and resource demands of each method. Our network, which relies heavily on attention mechanisms rather than extensive convolution layers, demonstrates a 35\% improvement in both parameter efficiency and memory usage. Moreover, our robust architecture and multi-view image utilization contribute to a 5.9\% enhancement in the $F_{\beta}$ score, reflecting improved performance and scalability.


\textbf{Visual Comparison}
\cref{fig:SOTAcompare} visually contrasts the performance of our method with selected state-of-the-art approaches from related fields. Our method consistently outperforms others in mirror detection. Existing techniques often misidentify mirror-like objects, such as wall frames or holes, and fail when objects are positioned directly in front of the mirror. They also struggle when the reflected scene closely resembles the surrounding environment. As illustrated in \cref{fig:SOTAcompare}, MVMD uniquely identifies details that other methods miss: it correctly detects the white bulbs in the right corner of the mirror in the first image, the books in front of the mirror in the fifth image, and the small mirror on the right side of the shelf in the seventh image, due to the effectiveness of our Inter-View and Intra-View Blocks. Additionally, MVMD successfully identifies the small angled section in the right corner of the eighth image, a feature highlighted by the Refinement network. Additional results for the MVMD method can be found in Appendix A.2.


\subsection{Ablation Study}
To validate our MVMD model, we conducted a series of ablation experiments. For input variations, we tested the model with only two images, as well as with three images captured from the same scene in a random order. For structural variations, we considered the Inter-Views Block and the initial mask decoder as the baseline configuration. We then assessed the model's performance by removing either the Intra-View Block or the Refinement Block, resulting in configurations that included the baseline plus one of these additional blocks.


\textbf{Effect of Suitable Inputs}
As illustrated in \cref{tab:ablation}, using only two images often fails to capture significant changes in the mirror reflections, leading to less accurate predictions. Furthermore, employing three randomly selected images from the same scene may result in the network failing to recognize the mirror area if the chosen views are not representative. This issue arises because differing perspectives of the same object can make it challenging for the network to learn object correspondences across images. In contrast, selecting three strategically chosen views provides a more comprehensive perspective, enhancing prediction accuracy and minimizing the risk of missing critical information.


\textbf{Impact of Individual Blocks}
As shown in \cref{tab:ablation}, the Intra-view Block plays a crucial role in learning object correspondences within the image, significantly enhancing the network's ability to predict mirror regions. Additionally, the Refinement Block markedly improves the clarity of mirror edges, especially in scenarios where objects are positioned in front of the mirror. This demonstrates the Refinement Block's essential role in enhancing edge precision and detail, making it a critical component for achieving high accuracy in mirror detection tasks.

\begin{table}[t]
    \centering
    \begin{tabular}{l c c c}
         \hline
         &  Parameter& Memory Usage&  $F_{\beta}$\\
         \hline
         MirrorNet\cite{mirrornet}&  121.77M& 464.50&  0.9279\\
         PDNet\cite{pdnet}&  80.54M& 307.24&  0.8855\\
         SANet\cite{sanet}&  103.07M& 403.73&  0.8659\\
         PMD\cite{PMD}&  147.66M& 563.28&  0.9259\\
         GlassNet\cite{glassnet}& 201.72M& 769.50& 0.9190\\
         ours&  \textbf{71.68M}& \textbf{283.06}&  \textbf{0.9442}\\
         \hline
    \end{tabular}
    \caption{Comparative analysis of computational complexity for MVMD. The best results are marked in bold for reference.}
    \label{tab:Complexity comparison}
\end{table}

\subsection{Attention visualization}
To gain a deeper understanding of our MVMD module, we visualize the attention heat maps. For each attention map, we compute the maximum value across all channels for each pixel, where yellow indicates high attention and dark blue indicates low attention.

\begin{figure}
    \centering
    \includegraphics[width=1\linewidth]{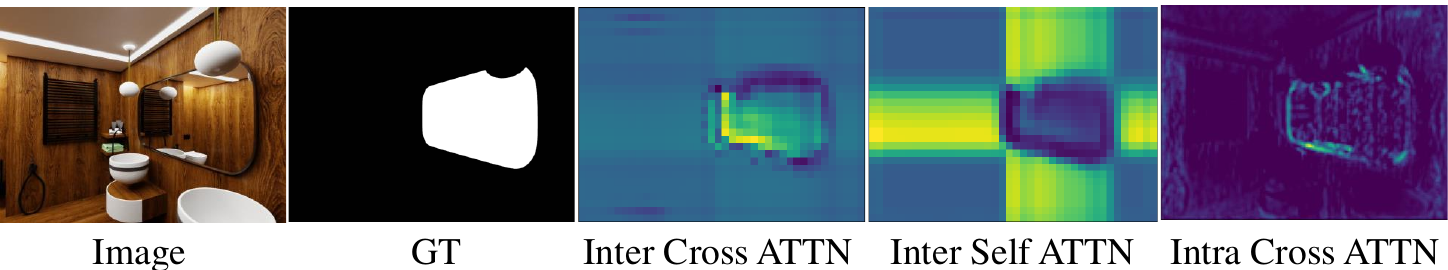}
    \caption{Visualization of attention heat maps. 'Inter Cross ATTN' represents the heat map for cross-attention in the Inter-Views Block, 'Inter Self ATTN' for self-attention in the same block, and 'Intra Cross ATTN' for cross-attention in the Intra-View Block.}
    \label{fig:attention}
    \vspace{-0.2cm}
\end{figure}

We first examine the cross-attention heat map between $I_1$ and $I_2$, and the self-attention heat map that follows the cross-attention in the Inter-Views Block. As shown in \cref{fig:attention}, the cross-attention highlights the left side of the mirror, where there is a noticeable difference in objects. The self-attention then focuses on the areas outside the mirror, further enhancing the learned information. For the Intra-View Block, the cross-attention heat map emphasizes the light bulb in the left corner and the wash basin at the bottom left, which correspond to objects outside the mirror and the mirror edge. For additional visualizations of attention and intermediate results, please refer to Appendix A.1.

\section{limitation and discussion}
\begin{figure}
    \centering
    \includegraphics[width=1\linewidth]{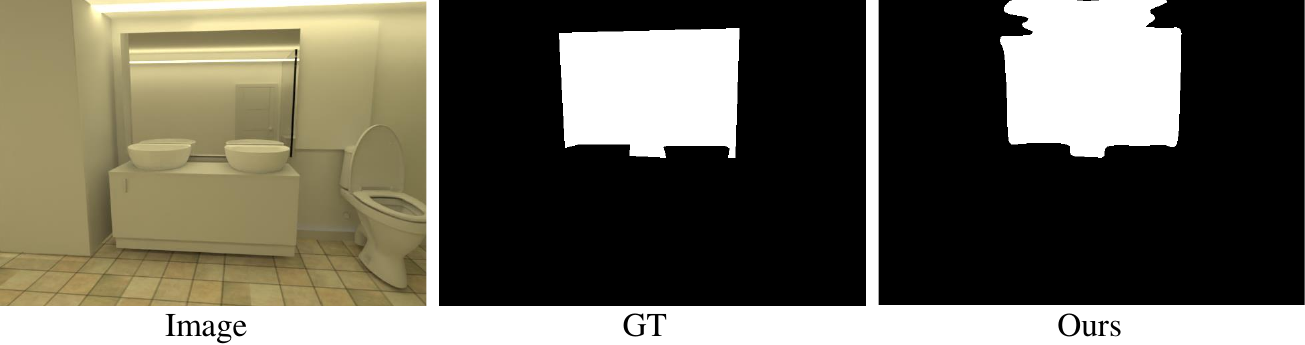}
    \caption{Example of failure cases due to similar reflections between mirror and surroundings.}
    \label{fig:Failure}
\end{figure}

In cases where objects within the mirror, such as walls or ceilings, lack distinctive features, the network may struggle to differentiate between various camera poses. This challenge arises because the absence of distinctive characteristics limits the network's ability to detect variations between different views. Additionally, when reflections in the mirror closely resemble real-world objects outside the mirror, the network may struggle to distinguish between the mirror and its surroundings, as illustrated in \cref{fig:Failure}.

Moreover, our method assumes that the images in each scene are ordered according to the sequence of camera movement, with the camera typically moving in a consistent direction. Consequently, the angular difference between the first and second images is usually smaller than that between the first and third images. If the images are not in the correct order, pre-processing is necessary to align them with our dataset's assumptions.

\begin{table}[t]
    \centering
    \begin{tabular}{l c c c c}
    \hline
         Method&  IOU&  MAE&  Accuracy& NMSE\\
    \hline
         baseline& 0.8597& 0.0153& 0.9847&0.1779\\
         b + refine&  0.8924&  0.0127&  0.9873& 0.1364\\
         b + intra&  0.8826&  0.0123&  0.9877& 0.1411\\
         $I_1$ + $I_2$&  0.9000&  0.0108&  0.9892& 0.1208\\
         rand 3 img&   0.9000&  0.0108&   0.9892&  0.1208\\
         full&  0.9019&  0.0106&  0.9894& 0.1183\\
    \hline
    \end{tabular}
    \caption{Quantitative comparison of ablation study results. 'b' denotes the baseline configuration, while 'rand 3 img' refers to using three images captured from the same scene in random order.}
    \label{tab:ablation}
\end{table}

\section{Conclusion}
In this paper, we introduce an innovative mirror detection method that utilizes multi-view images without requiring depth information. This technique is particularly effective for 3D reconstruction tasks and has the potential to enhance methods such as NeRF and 3D Gaussian Splatting. The MVMD network is designed with Inter-Views, Intra-View, and Refinement Blocks, each exploiting distinct characteristics of mirrors to enhance detection accuracy. Additionally, we provide a valuable Multi-View Mirror Detection (MVMD) dataset comprising 3,181 images from diverse scenes, which facilitates the training and evaluation of mirror detection models.

Our experimental results highlight the exceptional performance of MVMD, consistently surpassing state-of-the-art methods across various metrics. The method excels in distinguishing mirrors from similar objects and accurately detecting mirrors in challenging real-world conditions.

In summary, MVMD marks a significant advancement in Multi-View Mirror Detection, presenting a robust new tool for researchers and practitioners in computer vision and 3D reconstruction. By addressing the complex issue of mirror detection in multi-view scenarios, this work enhances the accuracy and reliability of 3D scene understanding.

\section{Acknowledgment} This research is partially supported by NSF grants CCF-2130688, and CNS-2107057.

{\small
\bibliographystyle{ieee_fullname}
\bibliography{egbib}
}

\end{document}




\appendix
\setcounter{figure}{0}
\setcounter{table}{0}

\section{Appendix/Supplemental material}

\subsection{Intermediate Results from MVMD Blocks}
To illustrate the functionality of each network component, we provide visualizations of attention heat maps and intermediate results. These visualizations demonstrate how the network processes input images, extracts information using attention mechanisms, and refines the mirror mask for accurate detection.

\begin{figure}[t]
    \centering
    \includegraphics[width=0.9\linewidth]{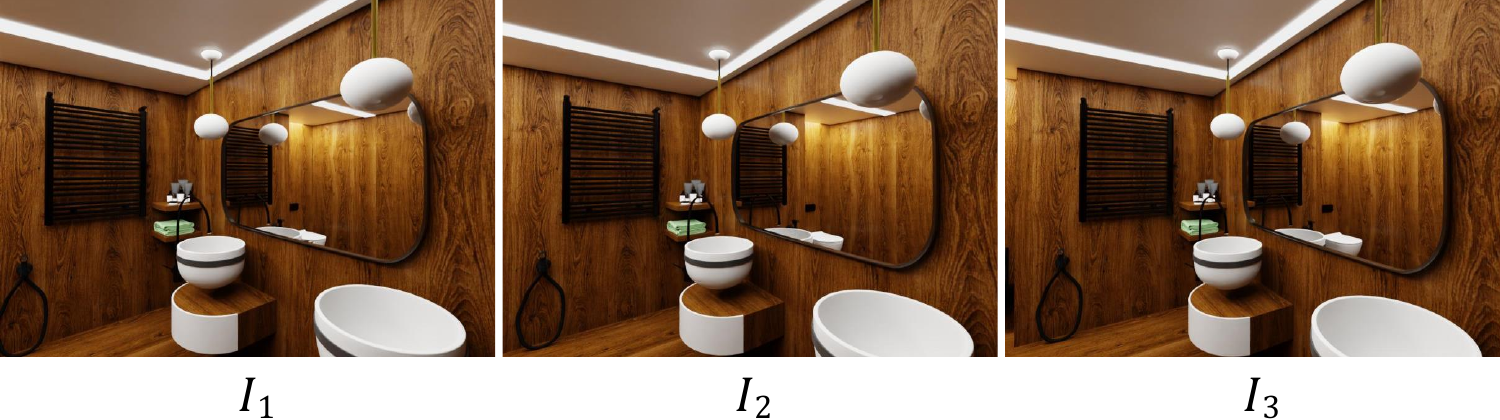}
    \caption{Example scene with three input images: $I_1$, $I_2$, and $I_3$ in an indoor bathroom.}
    \label{fig:3 input}
\end{figure}

As discussed in Section 4, our MVMD network takes three images from a single scene as inputs. \cref{fig:3 input} shows an example of an indoor bathroom scene with a large mirror and complex textures, which challenges mirror detection by making it difficult to distinguish the reflection area from the surrounding wall.

\begin{figure}[t]
    \centering
    \includegraphics[width=0.9\linewidth]{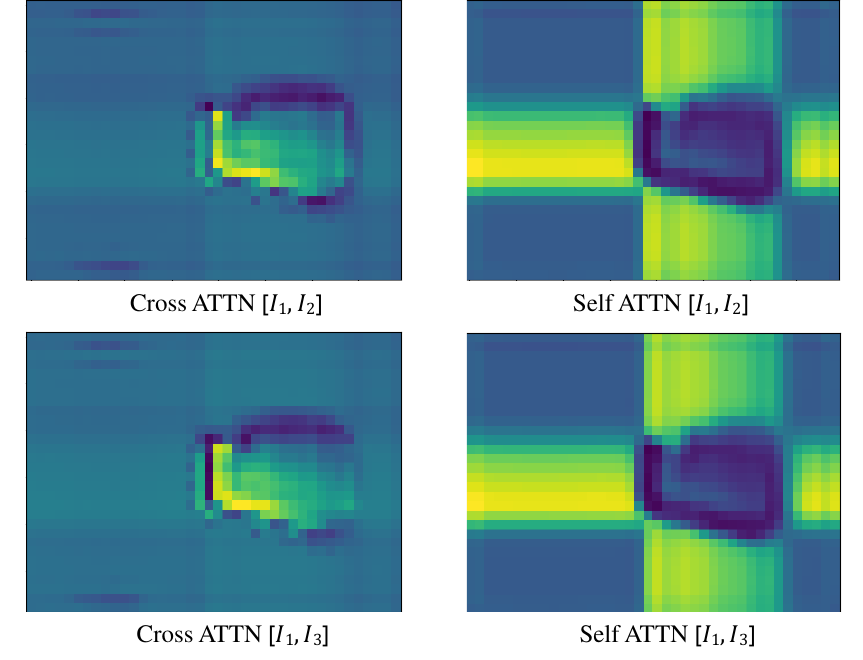}
    \caption{Visualization of the cross-attention and self-attention heat maps on [$I_1$, $I_2$] and [$I_1$, $I_3$] in the Inter-views Block.}
    \label{fig:inter heatl}
\end{figure}

\begin{figure}[t]
    \centering
    \includegraphics[width=0.9\linewidth]{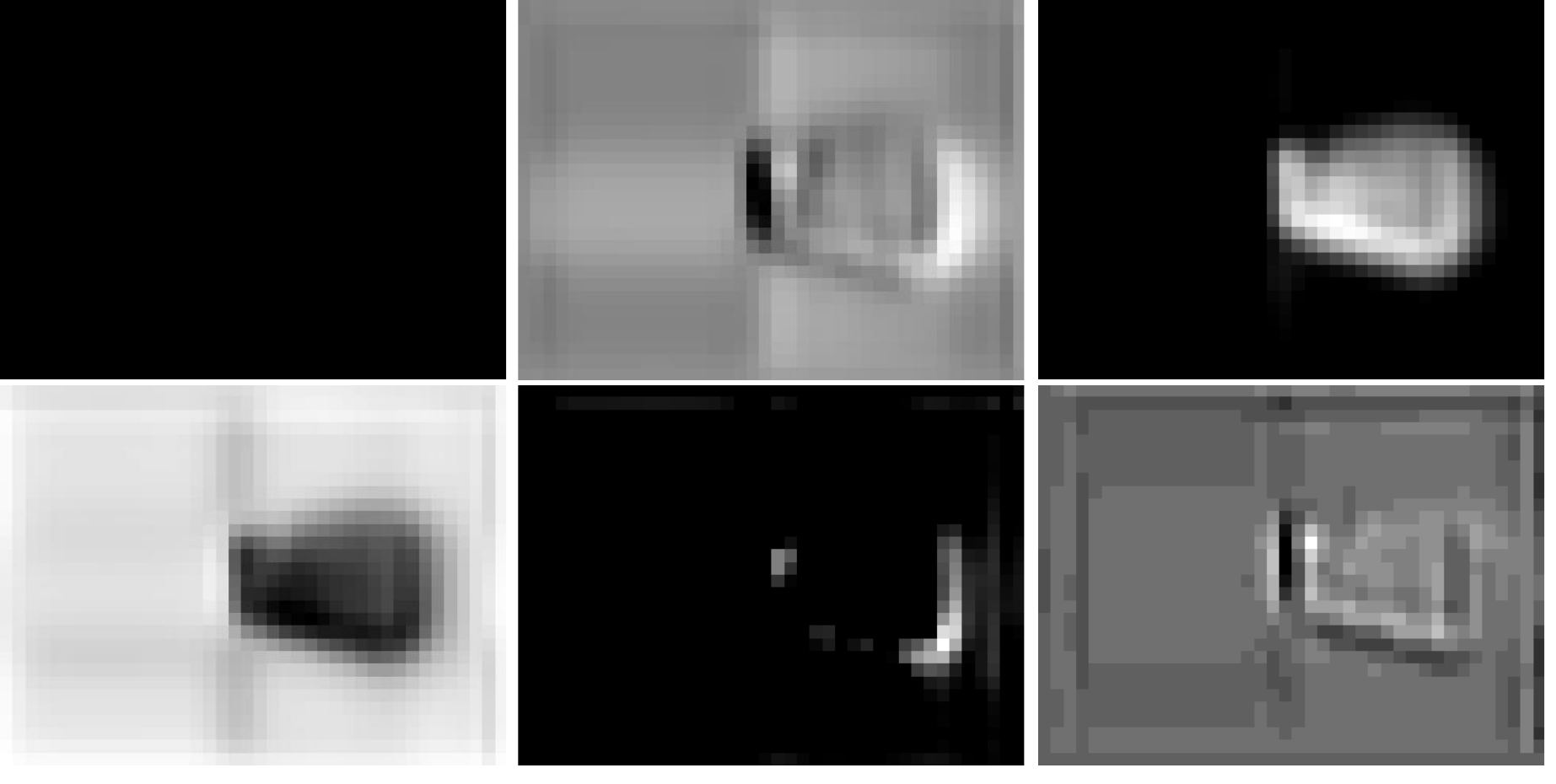}
    \caption{Visualization of selected channels from the output of the Inter-views Block.}
    \label{fig:inter out}
\end{figure}

The Inter-Views Block processes image pairs [$I_1$, $I_2$] and [$I_1$, $I_3$] through cross-attention and self-attention mechanisms. \cref{fig:inter heatl} visualizes the attention heat maps, revealing that cross-attention focuses on the left side of the mirror where reflection differences between [$I_1$, $I_2$] and [$I_1$, $I_3$] are evident. The differences in their cross-attention heatmaps indicate that each pair captures distinct mirror information due to viewpoint shifts and mirror reflections. Thus, using two pairs ensures that sufficient information is captured through these shifts. As discussed in Section 5.4, this shows the Inter-Views Block's ability to capture object shifts inside the mirror due to viewpoint changes. \cref{fig:inter out} depicts the output of the Inter-Views Block after channel attention. It shows that unimportant feature channels are filtered out (visible in the upper-left corner), while significant channels are retained, as demonstrated in the remaining figures.

\begin{figure}[t]
    \centering
    \includegraphics[width=0.9\linewidth]{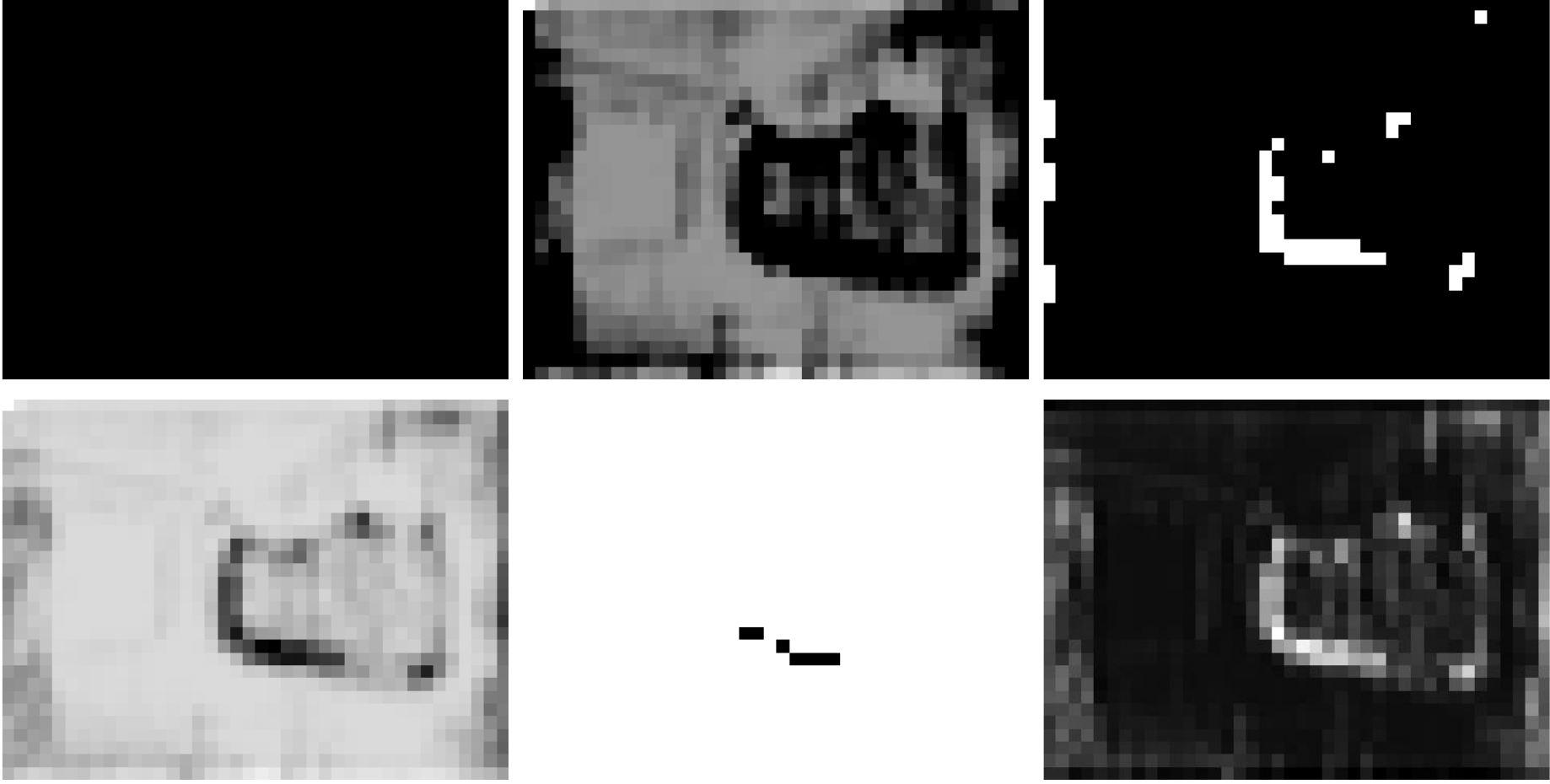}
    \caption{Visualization of selected channels from the output of the Intra-view Block.}
    \label{fig:intra out}
\end{figure}

As discussed in Section 5.4, the Intra-view Block captures the correspondence between objects inside and outside the mirror. \cref{fig:intra out} illustrates the output feature channels, highlighting how reflected objects inside the mirror match real objects outside. The same channel attention mechanism is applied to filter out less important channels while preserving key ones, ensuring precise mirror location information.

\begin{figure}[t]
    \centering
    \includegraphics[width=1\linewidth]{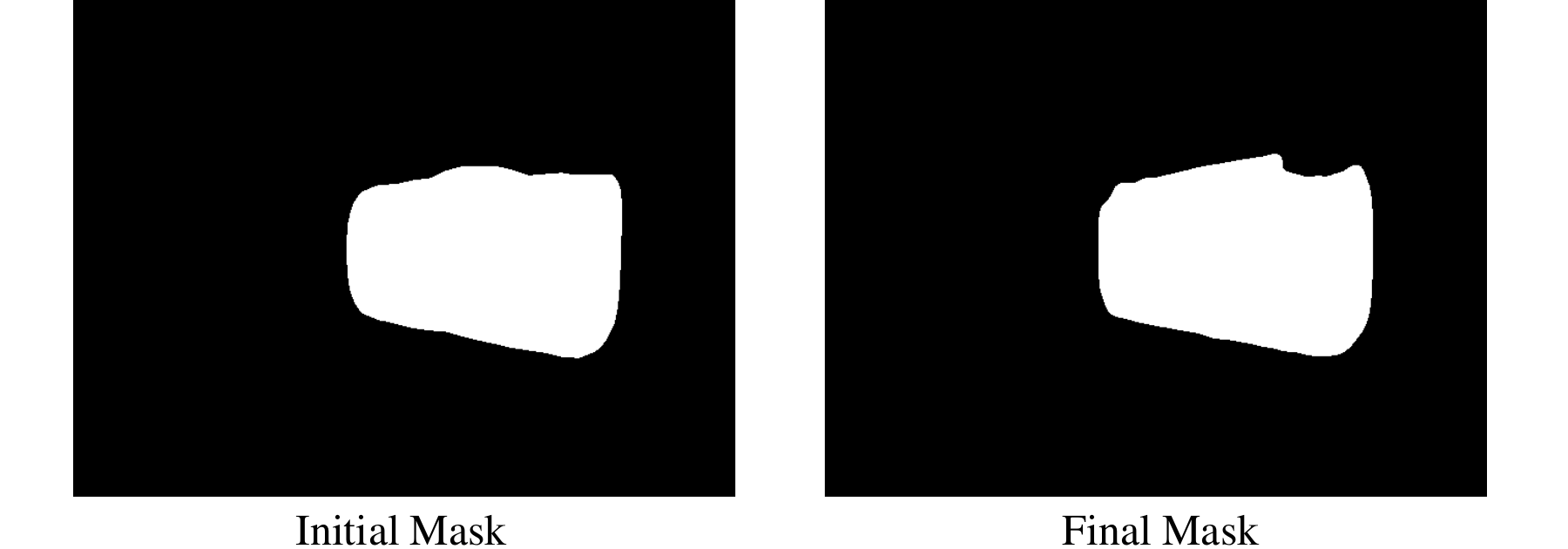}
    \caption{Visualization of the differences between the initial mask and final mask.}
    \label{fig:refine comapre}
\end{figure}

Finally, the Refinement Block enhances the edge definition of the initial mask generated by the Inter-View and Intra-view Blocks. \cref{fig:refine comapre} shows how it significantly improves mirror edge clarity and corrects misidentified mirror areas, particularly around objects in front of the mirror, such as the light bulbs at the top left and right.

\subsection{Additional MVMD Results}
\begin{figure*}[t]
    \centering
    \includegraphics[width=0.97\linewidth]{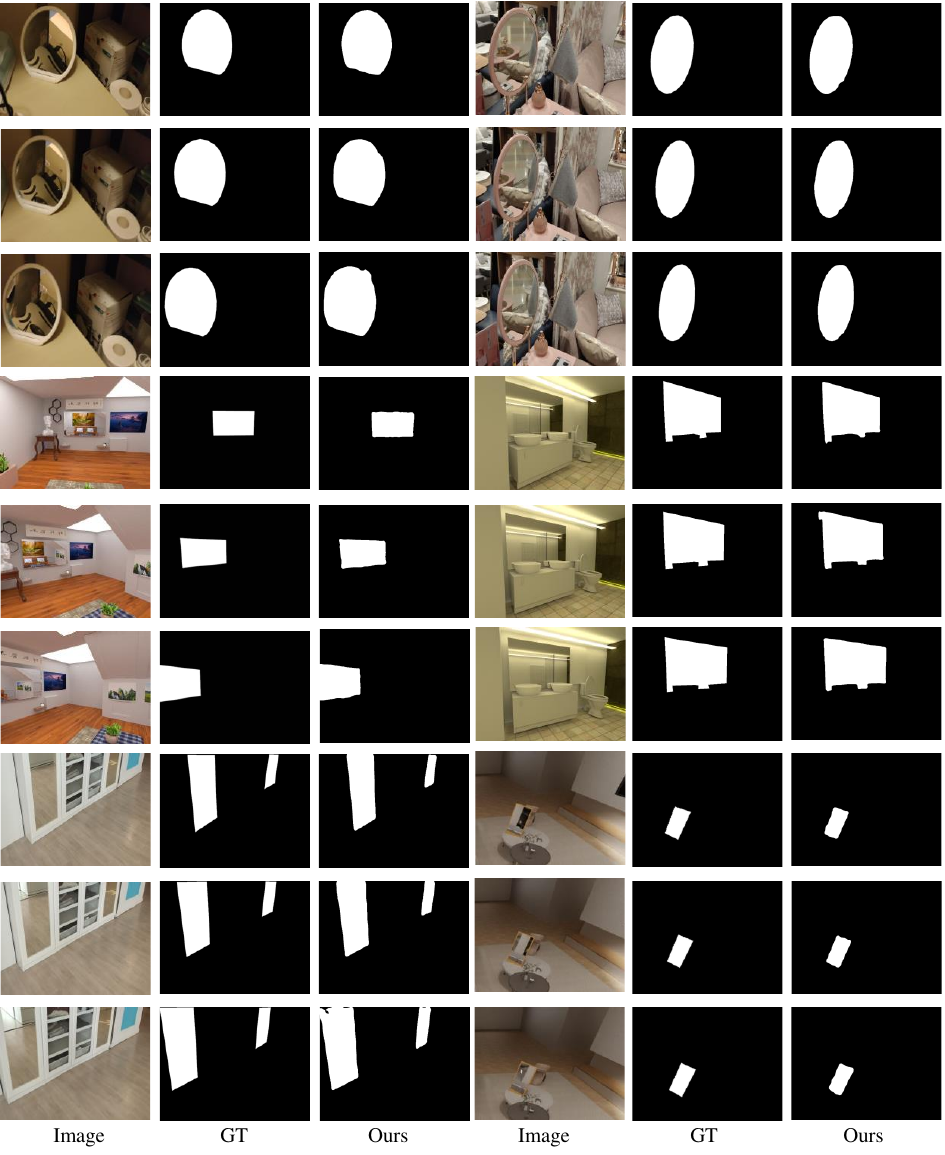}
    \caption{Additional MVMD network results on different scenes, featuring three viewpoints per example.}
    \label{fig:more_result}
\end{figure*}

To demonstrate the accuracy and robustness of the proposed MVMD module across a variety of scenes and its stability in detecting mirrors from multiple viewpoints within the same scene, we present additional inference results in \cref{fig:more_result}. For each scene, three images were selected from different angles of real-world environments. These results demonstrate the module’s high accuracy in detecting mirrors, even under challenging conditions such as indistinct edges and complex environments. Furthermore, they illustrate the module’s consistent performance across different perspectives, showcasing its ability to reliably identify and delineate mirrors regardless of viewpoint changes. This comprehensive evaluation underscores the effectiveness of the MVMD module in handling diverse and complex mirror detection scenarios.

\begin{figure*}[hpbt]
    \centering
    \includegraphics[width=0.97\linewidth]{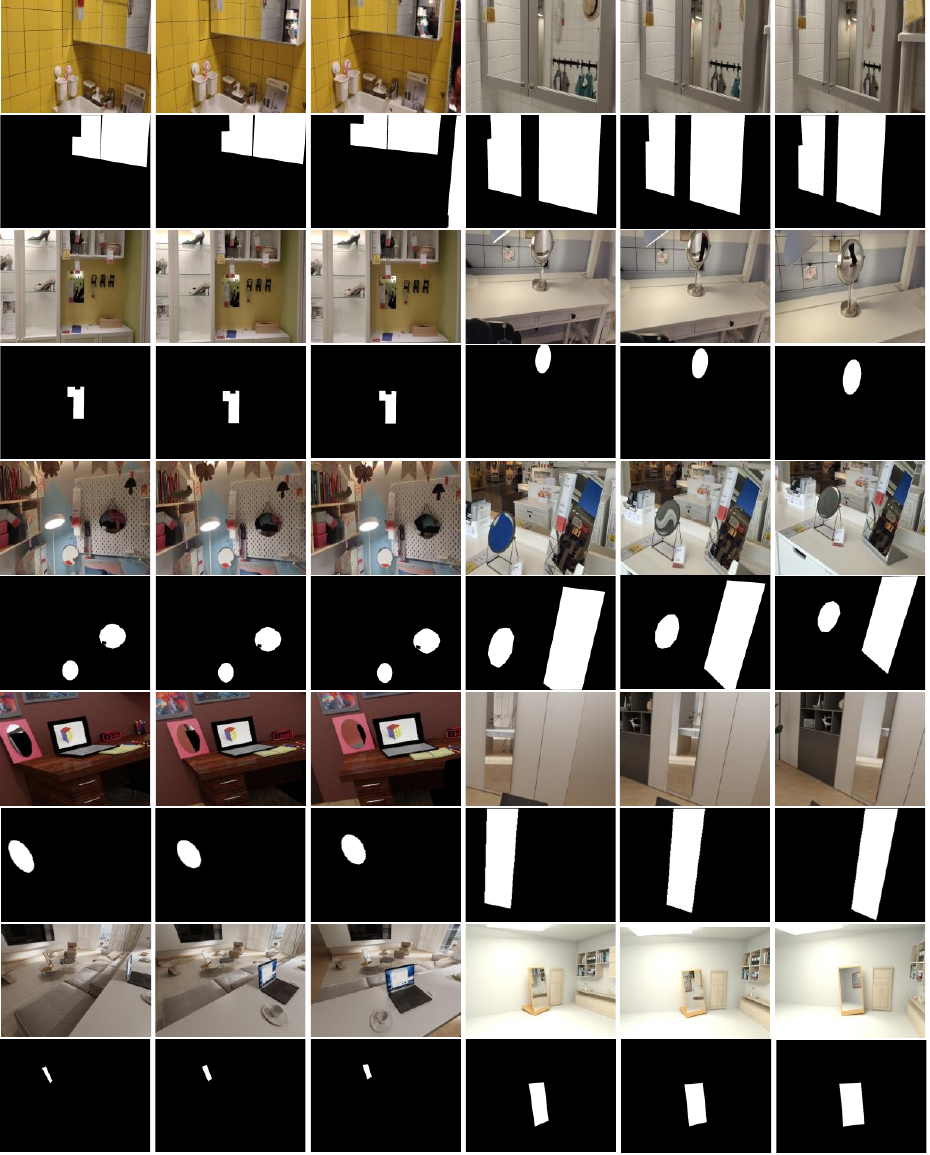}
        \caption{Additional examples from the proposed MVMD dataset.}
    \label{fig:more_dataset}
\end{figure*}

\subsection{Revisit MVMD Dataset}
To highlight the diversity of our MVMD dataset, particularly in terms of mirror shapes, sizes, locations, and quantities, we provide additional examples in \cref{fig:more_dataset}. These examples showcase a wide range of mirror configurations within various scenes, illustrating the dataset's capacity to cover different mirror types and placements. The variety in mirror attributes, such as large versus small mirrors, different shapes, and their diverse positions within the scene, underscores the comprehensive nature of the dataset. By including these varied examples, we demonstrate how the dataset captures the real-world complexity of mirror detection, which is crucial for training and evaluating the MVMD network under diverse conditions.

\subsection{From Mirror Mask to 3D Labels}
Various methods in 3D reconstruction use different types of image masks to modify scenes or enhance reconstruction quality based on specific tasks \cite{Weder2023Removing, mirzaei2023reference, Mirzaei2022SPInNeRFMS}. Image masks are particularly crucial in challenging areas such as mirrors, where they significantly improve accuracy \cite{liu2024mirrorgaussian, meng2024mirror3DGS, zeng2023mirror-nerf}. However, these masks are often created manually, a process that is labor-intensive and prone to errors, especially around complex edges. Such inaccuracies can result in sub-optimal reconstructions and artifacts in specific viewpoints.

To address these issues, we propose the MVMD network for automatically generating mirror masks, tailored to match the input requirements of 3D reconstruction tasks that use multi-view inputs. Our approach begins by generating a binary mask based on pixel-wise confidence scores from 2D images. This data-driven technique enhances the precision of mirror area identification, reducing manual errors and inconsistencies, and thus improving both reconstruction accuracy and efficiency.

Furthermore, to effectively utilize the automatically generated mirror mask, it can be mapped into 3D space using established labeling and voting algorithms, commonly employed in traditional 3D point cloud reconstructions. Labeling algorithms assign semantic categories to each point in the point cloud, aiding in scene classification and segmentation. Voting algorithms, such as Hough Voting \cite{qi2019deep}, RANSAC \cite{fischler1981random}, and kNN voting \cite{gou2011novel}, leverage global information from local features to enhance point cloud denoising, plane fitting, and object detection.

For NeRF (Neural Radiance Fields) \cite{mildenhall2021nerf} and 3D Gaussian Splatting (3D GS) \cite{kerbl3Dgaussians}, labeling and voting techniques can still be adapted, despite their different representations of 3D data compared to traditional point clouds. NeRF generates a continuous field using neural networks rather than explicit point clouds. Adaptations such as Semantic NeRF \cite{chen2022sem2nerf} extend the model to output semantic information for scene annotation. Similarly, the geometric structures produced by 3D GS can be converted into discrete point clouds \cite{kerbl3Dgaussians}, facilitating the use of traditional labeling techniques for classification and segmentation. Correspondingly, voting techniques can be applied indirectly to NeRF by first extracting point clouds from the NeRF output and then using these techniques to enhance segmentation or denoising. In the case of 3D GS, voting can be applied directly to point clouds or meshes obtained after the Structure from Motion (SfM) process, improving reconstruction accuracy and maintaining consistent object segmentation across multiple views.

By integrating these labeling and voting techniques, the proposed MVMD network effectively enhances the reconstruction accuracy of mirror regions in various 3D reconstruction tasks. It automatically generates mirror masks and maps them into 3D space, combining labeling with classification to precisely capture mirror regions and manage reflection effects. This approach reduces geometric errors and confusion caused by reflections, significantly improving the overall fidelity and reliability of 3D scene reconstruction. To further validate the effectiveness of the MVMD network, we plan to conduct a system-level integration and perform extensive experiments in future work to comprehensively assess its performance and potential applications in various complex scenarios. This will help to further optimize the model and ensure its reliability and effectiveness in practical applications.

\subsection{Extending to High-Reflectivity Regions}
In the real world, objects with high-reflective surfaces often exhibit characteristics similar to mirrors, which poses significant challenges for 3D reconstruction. High-reflectivity textures can mislead algorithms into interpreting reflections as part of the texture or as actual objects, leading to artifacts such as ghosting or errors depending on the viewing angle.

Directly applying mirror detection networks to identify all high-reflectivity areas remains challenging. Firstly, high-reflectivity surfaces exhibit greater variability compared to mirrors, as these surfaces are often uneven, leading to distortion or blurring in reflected content. Secondly, unlike mirrors with near-perfect reflections, high-reflectivity areas often display content influenced by the original texture and reflection coefficient, which can result in reflected external objects appearing as part of the surface. Finally, high-reflectivity areas may lack distinct edges, complicating the task of delineating actual object boundaries.

However, by leveraging the cross-attention and self-attention mechanisms of our MVMD network, there is potential to extend its capabilities to detect high-reflectivity regions. Specifically, the basic structure of our Inter-views Block can be retained to differentiate high-reflectivity areas from the background, as it is adept at identifying varying levels of reflection. For the Intra-view Block, we can adapt its design to handle the effects of original texture and reflection coefficients, enabling it to generate an initial mask for high-reflectivity regions. Finally, by enhancing the edge detection capabilities of the Refinement Block, we can improve the network's ability to distinguish texture-based edges within these high-reflectivity areas, refining the mask to better represent the actual boundaries of objects.

In future work, we plan to systematically evaluate and refine these extensions through extensive experimentation with various high-reflectivity scenarios. Our goal is to enhance the MVMD network’s robustness and accuracy in detecting and delineating high-reflectivity regions. By doing so, we aim to improve its applicability across a broader range of real-world environments, supporting next-generation 3D reconstruction applications.

{\small
\bibliographystyle{ieee_fullname}
\bibliography{egbib}
}